# Improving the Performance of Piecewise Linear Separation Incremental Algorithms for Practical Hardware Implementations

A. Chinea, J.M. Moreno, J. Madrenas, J. Cabestany

Universitat Politécnica de Catalunya, Departament d'Enginyeria Electrónica, c/ Gran Capitá s/n, 08034, Barcelona, Spain

**Abstract.** In this paper we shall review the common problems associated with Piecewise Linear Separation incremental algorithms. This kind of neural models yield poor performances when dealing with some classification problems, due to the evolving schemes used to construct the resulting networks. So as to avoid this undesirable behavior we shall propose a modification criterion. It is based upon the definition of a function which will provide information about the quality of the network growth process during the learning phase. This function is evaluated periodically as the network structure evolves, and will permit, as we shall show through exhaustive benchmarks, to considerably improve the performance (measured in terms of network complexity and generalization capabilities) offered by the networks generated by this incremental models.

## 1. INTRODUCTION

In the last few years a substantial effort in the field of Artificial Neural Networks' theory has been devoted to the study and development of incremental Neural Networks models [1], [2]. The most important feature of this kind of neural models is their ability to determine the proper network structure (i.e., the number of neurons and connections) to handle a particular task.

As was pointed out in [1], there are several types of incremental algorithms, and we shall concentrate on the Piecewise Linear Separation (PLS) models, due to the fact that they present a low compexity for both learning and recall phases, thus being well suited for VLSI implementations. PLS models are used mainly for classification tasks, and, starting from a network composed of just one neuron, they try to find in an incremental way the discriminant function able to separate the categories defined in the input space. This discriminant function is obtained by combining the linear discriminant functions associated with the perceptron-like units generated during the training process.

In this paper we shall first briefly review the methods which have been used for training the individual units generated by the PLS models, as well as the usual drawbacks posed by such algorithms. Then we shall present a novel modification criterion which can be applied to this incrementals models. This method will permit, given the desired generalization error as a starting parameter, to construct an appropiate network to meet such specification, allowing, at the same time a substantial reduction in the size of the generated network structures. Afterwards, we shall present a comparative and exhaustive simulation study on classification performance of the method proposed in this paper when is applied to a particular PLS model: The *Neural Trees* algorithm [3]. Finally, the conclusions and future work related to the proposed criterion will be outlined.

## 2. PROBLEM STATEMENT

Perceptron [4] and Pocket [4] are the most common training algorithms used for the units generated by this PLS incremental models. When the learning process is completed, each unit has the weight vector which vector which ideally yields the best correct classification rate in accordance with its input training set.

However, optimizing a function which accounts for the number of patterns correctly classified may impose an erroneus scheme for the incremental algorithm, so that, as stated in [6], a network structure of infinite size may be generated by the evolving process. Several methods have been already proposed in order to allow for a correct network evolving process [6], [7]. Among them, we have initially adopted the improvement method proposed in [6], which consist in running the Pocket algorithm with the modification that the two following conditions have to be met before a weight vector is stored as a new best weight vector: Both sides of the separating hyperplane defined by the weight vector are not empty, and the correct classification rate provided by this weight vector is larger than that provided by the best weight vector stored previously.

Nevertheless, this method generates quite complex network structures with rather poor generalization capabilities. Furthermore, the network structure depends on the order the inputs patterns are presented to the network during the training process. On the other hand, empirical observations carried out in artificial as well as in real databases have demonstrated that a large percentage of the units generated by the algorithm are used in the precise establishment of the boundaries used to separate the categories defined in the input space (causing therefore overfitting problems). This is due to the fact that PLS models present serious difficulties for solving problems in which there exists a high degree of overlapping between classes, and as a consequence produce a large amount of units trying to separate distributions of patterns very close to each other.

Taking into account the considerations stated above, we shall define a function which is calculated periodically during the training process and which detects when the network begins to have problems in determining the separation borders. This will permit to stop the growth of the network before it generates units which hardly provide information about the problem to be solved.

## 3. PROPOSED SOLUTION

Without loss of generality, let us consider a problem which consist in separatig two classes, denoted hereafter as class 0 and class 1. We define the class-i (i=0,1) centroid, $c_i$, as the vector whose components are obtained by calculating the mean value from the components of the vectors which represent the patterns belonging to this class.

Let us define the following functions:

$$J_1 = \frac{1}{N_{10} + N_{11}} \left[ \sum_{i=1}^{N_{10}} \sum_{j=1}^{d} \frac{1}{\left| x_{ij}^{10} - c_j^{11} \right|} + \sum_{i=1}^{N_{11}} \sum_{j=1}^{d} \frac{1}{\left| x_{ij}^{11} - c_j^{10} \right|} \right] \qquad (1)$$

$$J_0 = \frac{1}{N_{00} + N_{01}} \left[ \sum_{i=1}^{N_{00}} \sum_{j=1}^{d} \frac{1}{\left|x_{ij}^{00} - c_j^{01}\right|} + \sum_{i=1}^{N_{01}} \sum_{j=1}^{d} \frac{1}{\left|x_{ij}^{01} - c_j^{00}\right|} \right] \quad (2)$$

$$J_c = J_0 + J_1 \quad (3)$$

Where:

- d: Dimension of the input data space.
- $N_{kl}$ : Number of patterns belonging to class k the network classifies as belonging to class l (k,l=0,1).
- $x_{ij}^{kl}$: j-th component of the i-th pattern classified by the network as class k but belonging to class l.
- $c_{ij}^{kl}$: j-th component of the centroid determined for the patterns which belong to the class k and are classified by the network as class l.

As it can be deduced, this function will present maximum values when the network begins to have problems in determining the separation borders between the two categories. This is the reason for calculating in the auxiliary functions $J_1$ and $J_0$ the inverse of the distance from the components of the patterns to the components of the centroids, so as to emphasize this fact. On the other hand, these auxiliary functions are normalized, so that they do not depend on the number of patterns. Moreover, in the case any of the $c^{kl}$ centroids do not exist, the corresponding function $J_k$ will have value zero.

The initial value of $J_c$, $J_{ci}$, before the learning process is started, can be thus defined, using the same notation as for (1)-(3), as follows:

$$J_{ci} = \frac{1}{N_0 + N_1} \left[ \sum_{i=1}^{N_0} \sum_{j=1}^{d} \frac{1}{\left|x_{ij}^0 - c_j^1\right|} + \sum_{i=1}^{N_1} \sum_{j=1}^{d} \frac{1}{\left|x_{ij}^1 - c_j^0\right|} \right] \quad (4)$$

Figure 1 depicts the evolution of the $J_c$ function when trying to solve the classification problem stated in the phoneme database, provided by the ROARS Esprit Project. The aim of this database is to distinguish between nasal and oral vowels (therefore, two different categories are defined in the input space) coming from 1809 isolated syllabes. Each vector constituting this database is characterized by five features, corresponding to the first harmonics, normalized by the total energy. The database is composed of 5404 vectors, 3818 of class 0 (nasal vowel) and 1586 of class 1 (oral vowel).

The algorithm used to train the network is the Neural Trees algorithm [3], which belongs to the category of PLS models. This algorithm tries to solve a certain classification task by dividing iteratively the initial problem in subsequent reduced versions, which are used as training sets for the units generated during the network construction process. As a consequence, this algorithm produces finally a network structure which resembles a binary decision tree. In this way, Fig.1 represents the value of the $J_c$ function for each level of the tree generated by the incremental algorithm. The unit training principle has been the Pocket algorithm with the modification criterion mentioned previously. The number of iterations (i.e., the number of weights updates) for each unit was set to 10000.

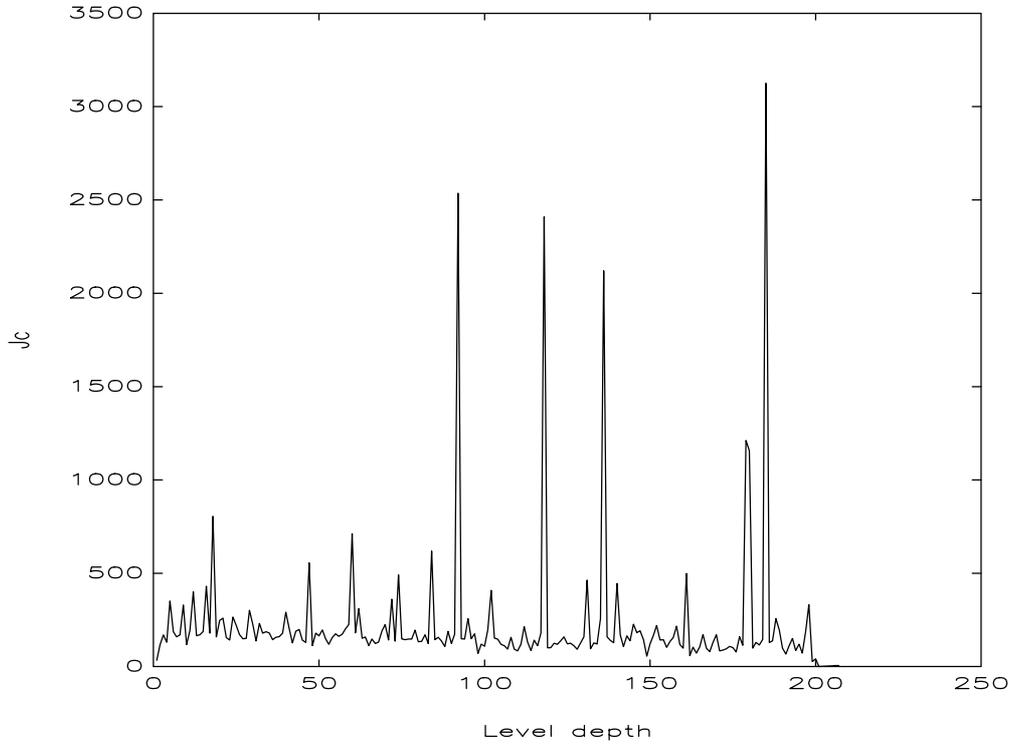

**Fig.1.** Evolution of the $J_c$ function for the *phoneme* database

The evolution of the $J_c$ function depicted in Fig.1 suggest a stopping criterion for the network growth process. It consist in freezing the network construction procedure when a relative peak (with respect to a given reference) of the $J_c$ function arises. We have set this reference as the value for the $J_c$ function before the training process is started. Thus, if we further use some penalty/merit functions in order to maximize the quotient generalization / network complexity, we shall improve the overall performance provided by the network constructed by PLS incremental models.

A suitable penalty/merit function, $I$, could be defined as follows:

$$I = \begin{cases} \left[\dfrac{G}{1+\left(\dfrac{G-G_{max}}{G_{max}}\right)^2}\right]^2 \dfrac{1}{1+\left(\dfrac{C}{C_{max}}\right)} & G \leq G_{max} \\ \\ \dfrac{G^2}{1+\left(\dfrac{C}{C_{max}}\right)} & G > G_{max} \end{cases} \quad (5)$$

Where *G* is the generalization given by the network (i.e, the correct classification rate provided for the test set) and *C* is the complexity of the network (measured as number of units). The constant parameters $G_{max}$ and $C_{max}$ can be determined using the theoretical results obtained by Baumm and Haussler [8] concerning the probability of poor generalization. More specifically, they showed that at least about $W/\varepsilon$ training examples are needed to obtain a generalization error less than ε, being *W* the number of weights in the network. In the case of the Neural Trees algorithm the previously described lower bound on the number of patterns leads to the following expression:

$$C_{max} = \frac{p\varepsilon}{d+1} \qquad (6)$$

Where p is the number of patterns on the training set, ε the generalization error and d the dimension of the input space. Thus, we have completely characterized expression (5) since we have related the maximum complexity of the network $C_{max}$ with the generalization error $\varepsilon$. Hence, once the value of the generalization error is fixed, the parameters $G_{max}$ (since $G_{max}$ = 1 - ε) and $C_{max}$ in function I are determided automatically.
setting the value of the generalization error we will have automatically determined the parameters $G_{max}$ and $C_{max}$ in function *I*.

Taking into account the considerations stated above we can summarize the proposed network evolving scheme as follows:

1. Given a desired generalization error calculate the value of the function $J_c$ for the entire distribution.
2. Start the network evolving scheme imposed by the particular PLS model. During the network construction process, evaluate periodically the function $J_c$ in order to detect peaks of amplitud λ times greater than the initial value.
3. If the function $J_c$ presents a peak value, then calculate the current value of the function I. If this value is greater than that calculated in the previous maximum, it is updated and the network construction process is still allowed. Otherwise, the network evolution is stopped at this point.

It is important to note that the election of the parameter λ may be critical. If we choose a small value for λ the network growth process will be stopped to early, even when the network size will not be enough for providing a satisfactory correct classification rate, leading to misclassifications and a resulting network structure with rather poor generalization capabilities.

On the other hand, if we choose an arbitrary large value for the parameter λ the complexity of the generated network structures will increase due to the fact that maximum values given by the function $J_c$ never reach the value given by the parameter λ. Furthermore, even in case we could choose an optimal value for λ the standard deviation measured in terms of the complexity of the structures generated during the training phase, would be quite large as a consequence of the behavior of the function $J_c$. The maximum values presented by the function $J_c$ are highly dependents on the particular evolution followed by the network during the learning phase, specifically on the order the training patterns are presented to the network. Therefore, in order to avoid the problems previously addressed we finally defined the parameter λ as a linear decreasing function over the complexity (number of units) followed by the network during the course of training.

Expression (7) shows the above mentioned relationship between the parameter λ and the network complexity $C$. It is simple to see how the parameter λ will take large values while the complexity of the network remains small, decreasing its value at the extent network complexity grows up. Thus, the proposed evolving scheme will generate structures with a complexity around the optimal value given by $C_{max}$.

$$\lambda = \begin{cases} C_{max}\left(1 - \dfrac{C}{C_{max}}\right) & C \leq C_{max} \\ 0 & C > C_{max} \end{cases} \qquad (7)$$

In the next section we shall present a comparative simulation study on classification performance of the Neural Trees algorithm and the Neural Trees algorithm with the proposed method.

## 4. EXPERIMENTAL RESULTS

The classification tasks used to check the proposed modification criterion are contained in the following artificial and real databases:

- *gauss2*, *gauss4*, *gauss6* and *gauss8*: These four databases are composed by 5000 vectors belonging to two normal distributed classes (2500 vectors for each category) in dimensions 2, 4, 6 and 8, respectively. Both distributions have the same mean but different variance.
- *rectangular*: It is composed of 2500 bidimensional vectors, 1248 belonging to class 0, and 1252 belonging to class 1, which are distributed following bidimensional distributions in the square (0,0)-(1,1) with overlapping.
- *clouds*: This database consists of 5000 vectors belonging to two different classes, with 2500 vectors in each class. The first class is obtained by the sum of three different gaussian distributions, while the second class corresponds to a single normal distribution.
- *phoneme*: The main features of this database have been explained previously in section 3.

The experiments were carried out using the leave-k-out cross-validation procedure [9]. In this way, the original database is divided in ten equal sized parts, and then the network is trained with nine parts and tested with the remaining part, so that a total of ten training-test sets are obtained for each database. This process is repeated six times for each partition, so that finally a total amount of sixty evolving processes are performed for each database.

Table 1 shows the results, indicated as mean number of units generated and generalization percentage (together with the corresponding standard deviation of both values) provided by the Neural Trees algorithm ( which evolves a network structure providing a 100 % correct classification rate for the training set) for the classification tasks stated previously.

| Problem | Number of units | Std. dev. | Generalization (%) | Std. dev. |
|---|---|---|---|---|
| *gauss2* | 1748.0 | 24.26 | 63.66 | 1.91 |
| *gauss4* | 1117.1 | 21.74 | 68.6 | 2.09 |
| *gauss6* | 891.0 | 25.29 | 72.2 | 2.14 |
| *gauss8* | 752.77 | 26.05 | 73.78 | 1.95 |
| *rectangular* | 626.43 | 12.77 | 76.62 | 2.34 |
| *clouds* | 762.13 | 19.84 | 84.6 | 2.55 |
| *phoneme* | 744.0 | 22.4 | 84.9 | 1.51 |

**Table 1.** Results for the *Neural Trees* algorithm

Table 2 reproduces the results provided by the Neural Trees algorithm when the network evolving process is modified by the criterion proposed in this paper. In our experiments, the desired generalization error was set to that given by the theoretical Bayes limit. In the case of the phoneme database we tested differents values for the generalization error since Bayesian limit was not known.

As can be deduced from the comparison of these tables, the results provided by the proposed method not only represents a substantial reduction on the number of units generated by the algorithm, but also a meaningful improvement on the generalization capability. However, simulation results corresponding to the clouds and the gaussian database (except for the gauss2 problem) illustrate the difficulty in obtaining generalization errors close to the Bayesian limit. For the case of the gaussian database this behavior can be explained due to inherent sparseness of high dimensional training data (known as the curse of dimensionality).

To have a better understanding about this effect it is important to note that for the gaussian database the number of training patterns is the same in all dimensions. Hence, from statistical point of view it does not exists an amount of enough training samples in order to estimate a probability density distribution. Moreover, linear smoothers (like the PLS models) generally have insufficient data in high dimensional spaces to reliably estimate a probability density distribution.

| Problem | Number of units | Std. dev. | Generalization(%) | Std. dev. | Error (%) |
|---|---|---|---|---|---|
| *gauss2* | 428.56 | 11.26 | 71.36 | 1.94 | 26.37 |
| *gauss4* | 200.12 | 11.25 | 76.49 | 1.98 | 17.64 |
| *gauss6* | 120.8 | 10.35 | 77.89 | 1.89 | 12.44 |
| *gauss8* | 80.31 | 9.46 | 76.56 | 1.9 | 9.0 |
| *rectangular* | 122.33 | 5.26 | 82.55 | 2.2 | 15.51 |
| *clouds* | 157.2 | 6.22 | 78.25 | 1.89 | 9.66 |
| *phoneme* | 133.36 | 6.16 | 83.72 | 1.39 | 15.0 |
| *phoneme* | 97.53 | 8.38 | 83.23 | 1.5 | 10.0 |
| *phoneme* | 61.68 | 7.43 | 83.05 | 1.28 | 5.0 |

**Table 2.** Results for the *Neural Trees* algorithm modified with the proposed criterion

In the case of the clouds database not only the previously commented difficulties related to the Bayes limit can be observed, but also a substantial looseness of the generalization error with respect to that given in table 1. At this point, it is helpful to note that the lower bound given in the previous section (see expression (6)) was made under the assumption of a network with a large number weights. Nevertheless, this fact constitutes the main reason of such discouraging results, since working conditions of the mentioned hypothesis fail.

A natural way to overcome this limitation is to use the upper bound on the number of patterns given by Baum and Haussler. More specifically, they showed that if the error on the training set was less than $\varepsilon/2$ at most of the order of $(W/\varepsilon)\log(M/\varepsilon)$ examples are needed to obtain a generalization error less than $\varepsilon$. Where $W$ is the number of weights on the network and $M$ the number of threshold units. For the case of the Neural Trees algorithm the following expression is straightforward to follow:

$$p = \frac{C_{max}(d+1)}{\varepsilon} \log\left(\frac{C_{max}}{\varepsilon}\right) \qquad (8)$$

Where p is the number of patterns, d the dimension of the input space and $C_{max}$ is the complexity of the network. However, the value of the parameter $C_{max}$ is very difficult to compute due to the non-linearity of the equation derived above.

A heuristic for avoiding such a problem can be summarized as follows:

1. Compute the parameter $C_{max}$ from expression (6)
2. Given the value of $C_{max}$ compute $p$ from expression (8).
3. Re-compute $C_{max}$ from expression (6) ussing the new value of $p$.

Table 3 shows the results provided by the Neural Trees algorithm when the network evolving process is modified by the criterion expounded in section 3 together with the previously described heuristic procedure.

From the observation of these results, it is clear to see how the proposed heuristic overcome the aforementioned difficulties. On the other hand, for the clouds database, as was expected, it is appreciated a meaningful increase of the generalization capability.

| Problem | Number of units | Std. dev. | Generalization(%) | Std. dev. | Error (%) |
|---|---|---|---|---|---|
| *gauss2* | 1259.5 | 4.26 | 67.32 | 2.13 | 26.37 |
| *gauss4* | 486.13 | 6.3 | 75.87 | 1.77 | 17.64 |
| *gauss6* | 255.98 | 11.53 | 77.88 | 1.7 | 12.44 |
| *gauss8* | 161.46 | 10.57 | 77.71 | 2.2 | 9.0 |
| *rectangular* | 335.78 | 8.51 | 80.1 | 2.08 | 15.51 |
| *clouds* | 486.31 | 9.82 | 84.0 | 2.09 | 9.66 |
| *phoneme* | 353.15 | 15.42 | 84.54 | 1.52 | 15.0 |
| *phoneme* | 239.25 | 4.25 | 84.2 | 1.5 | 10.0 |
| *phoneme* | 129.73 | 6.25 | 83.91 | 1.38 | 5.0 |

**Table 2.** Results for the *Neural Trees* algorithm modified with the proposed criterion and the heuristic

As a consequence of the results presented previously, two main conclusions can be derived:

- The proposed criterion is able to reduce the network complexity generated by the incremental algorithm, thus facilitating an eventual hardware implementation of the classifier evolved by the PLS incremental model. Furthermore, the standard deviation of the number of units generated is considerably reduced, minimizing the influence of the vector presentation order during the training phase.
- The generalization performance provided by the final network structure is held or even improved, as was expected for less complex networks.

## 5. CONCLUSIONS AND FUTURE WORK

We have shown in this paper that the linear discriminant solutions provided by the usual training algorithms are not always useful for evolving the network structures generated by PLS incremental models.

After reviewing some modification proposal for these training algorithms and the problems posed by the network evolving scheme associated with PLS models, we have introduced a novel approach for improving the performance of these models. This modification criterion is able to stop the network construction process when no further improvement is obtained by adding new units to the network. Furthermore, the method permits to construct automatically the proper network structure for a given classification task with only one input parameter, the generalization capability expected for the resulting classifier.

As a consequence, the proposed method produces very compact network structures for a given problem to be handled, thus facilitating an eventual hardware implementation (or software emulation). Together with the expected improvement in the generalization capabilities of the resulting networks, this method is able to reduce the influence of the vector presentation order during the training phase of the network evolving process.

Our current work is concentrated in comparing the proposed method with some criterions proposed recently [10] and aimed also at selecting the proper network structure able to handle a given classification task. We contemplate also the possibility to apply this method to incremental models aimed at solving regression tasks.